\documentclass[lettersize, journal]{IEEEtran}
\usepackage[T1]{fontenc}

\usepackage{amsmath,amssymb,amsfonts}
\usepackage{graphicx}
\usepackage{pifont}
\usepackage{filecontents}

\usepackage{amsthm}
\usepackage{xcolor}
\usepackage[figuresright]{rotating}

\usepackage{graphicx}
\graphicspath{{/}{fig/}}

\usepackage{array}
\usepackage{textcomp}
\usepackage{xcolor}
\usepackage{multirow}
\usepackage{booktabs}

\usepackage{mathtools}
\usepackage{float}

\usepackage{pgfplots}
\pgfplotsset{compat=1.18}
\usepgfplotslibrary{groupplots}
\usepackage{pgfplotstable}
\usepackage{svg}

\usepgfplotslibrary{groupplots}
\usepgfplotslibrary{statistics}

\usepackage{caption}
\usepackage{subcaption}
\usepackage{orcidlink}

\usepackage{multirow,tabularx}
\usepackage{hyperref}
\usepackage{flushend}
\usepackage{algorithmic}
\usepackage[vlined, ruled, shortend]{algorithm2e}

\definecolor{AviaColor}{RGB}{228,26,28}
\definecolor{Mid360Color}{RGB}{55,126,184}
\definecolor{OusterColor}{RGB}{77,175,74}

\usepackage{enumitem}
\usepackage{underscore}

\newlist{modlist}{enumerate}{1}
\setlist[modlist]{label=\textbf{(\arabic*)}, leftmargin=*, itemsep=2pt, topsep=2pt}

\newlength\figureheight
\newlength\figurewidth
\SetAlCapNameFnt{\footnotesize}
\SetAlCapFnt{\footnotesize}
\title{A Sensor-Aware Phenomenological Framework for Lidar Degradation Simulation and SLAM Robustness Evaluation}

\author{%
  Doumegna~Mawuto~Koudjo~Felix\,\orcidlink{0009-0008-8469-6573},
  Xianjia~Yu\,\orcidlink{0000-0002-9042-3730}\,
  Zhuo~Zou\,\orcidlink{0000-0002-8546-1329},
  Tomi~Westerlund\,\orcidlink{0000-0002-1793-2694}%

\thanks{This research is supported by the Research Council of Finland's Digital Waters (DIWA) flagship (Grant No. 359247).}
\thanks{All authors are with \href{https://tiers.utu.fi}{Turku Intelligent Embedded and Robotic Systems (TIERS) Lab}, University of Turku, Turku, Finland.(e-mail:\{mawuto.k.doumegna, xianjia.yu, tovewe\}@utu.fi.)}
\thanks{Doumegna Mawuto Koudjo Felix  and Zhuo Zou are also with School of Information Science and Technology, Fudan University, Shanghai, China.(e-mail:\{23210720352, zhuo\}@m.fudan.edu.cn.)}
 
}

\begin{document}

\maketitle

\thispagestyle{empty}
\pagestyle{empty}

\begin{abstract}
Lidar-based SLAM systems are highly sensitive to adverse conditions such as occlusion, noise, and field-of-view (FoV) degradation, yet existing robustness evaluation methods either lack physical grounding or do not capture sensor-specific behavior. This paper presents a sensor-aware, phenomenological framework for simulating interpretable lidar degradations directly on real point clouds, enabling controlled and reproducible SLAM stress testing. Unlike image-derived corruption benchmarks (e.g., SemanticKITTI-C) or simulation-only approaches (e.g., lidarsim), the proposed system preserves per-point geometry, intensity, and temporal structure while applying structured dropout, FoV reduction, Gaussian noise, occlusion masking, sparsification, and motion distortion.
The framework features autonomous topic and sensor detection, modular configuration with four severity tiers (\textit{light}–\textit{extreme}), and real-time performance ($<20$\,ms per frame) compatible with ROS workflows. Experimental validation across three lidar architectures and five state-of-the-art SLAM systems reveals distinct robustness patterns shaped by sensor design and environmental context. The open-source implementation provides a practical foundation for benchmarking lidar-based SLAM under physically meaningful degradation scenarios.
\end{abstract}

\IEEEpeerreviewmaketitle


\section{Introduction}\label{sec:introduction}

Lidar sensors play a vital role in robotic perception tasks, yet their performance degrades under adverse conditions such as fog, dust, occlusion, and dynamic motion, leading to 
and reduced SLAM reliability~\cite{pritzl2025degradation}. Prior studies have characterized these effects. Bijelic et al.~\cite{9157107} analyzed fog-induced attenuation, Carballo et al.~\cite{ZHANG2023146} reviewed degradation taxonomies, and Wu et al.~\cite{10898027} introduced an uncertainty-aware Lidar–IMU fusion approach that models measurement noise to improve odometry robustness, highlighting the importance of understanding lidar degradation. Recent surveys~\cite{10186539} further emphasize the need for physically interpretable models grounded in real sensor behavior. However, these works do not provide modular, sensor-specific simulation frameworks tailored for robustness evaluation in SLAM. 

Robustness benchmarks such as SemanticKITTI-C, nuScenes-C,  KITTI-C, and TIERS Dataset~\cite{10286105,10205484,DBLP:journals/corr/abs-2409-10824,sier2023benchmark} evaluate perception under unified 3D corruptions. While valuable for downstream tasks, these benchmarks apply generic perturbations that lack fine-grained modeling of different lidar hardware types. Augmentation methods such as PolarMix~\cite{xiao2022polarmix}, Real3D-Aug~\cite{sebek2023real3daug}, and RealAug~\cite{zhan2023realaugrealisticscenesynthesis} target neural network training and often rely on artificial or learned perturbations that do not reflect physical sensor behavior. As a result, they provide limited insights into SLAM or lidar-inertial odometry (LIO) performance under realistic lidar degradations.

To address this gap, we propose a sensor-aware, phenomenological framework that applies physically interpretable degradations directly to raw lidar measurements, enabling controlled, reproducible evaluation of SLAM robustness. Our design builds on findings in~\cite{ZHANG2023146,10898027,9157107,10186539}, emphasizing interpretable modeling of noise, occlusion, FoV reduction, motion distortion, and other environment-induced artifacts. Unlike augmentation pipelines designed for data-driven training, our objective is to simulate realistic perturbations rather than improve model generalization, enabling structured benchmarking under adverse conditions. This work extends our previous dataset~\cite{11248862}, which characterized Livox Avia, Livox Mid-360, and Ouster lidars under nominal conditions, by introducing controlled degradation mechanisms applicable across diverse sensor architectures.

Our contributions are summarized as follows:
\begin{itemize}[leftmargin=*]
\item \textbf{Sensor-aware degradation modeling}: capturing scanning-pattern differences across solid-state (Livox Avia, Mid-360) and spinning (Ouster) lidars.
\item \textbf{Modular and reproducible configuration}: parameterized degradation scenarios with four severity tiers (\textit{light}–\textit{extreme}), including dropout, FoV reduction, noise, occlusion, sparsification, and motion distortion.
\item \textbf{Real-time ROS implementation}: hybrid C++/Python framework with autonomous topic and sensor detection, integrated visualization, and $<20$\,ms processing latency.
\item \textbf{Systematic SLAM robustness evaluation}: enabling physically grounded comparison across heterogeneous sensors and state-of-the-art SLAM under adverse conditions.
\end{itemize}

\noindent\textbf{Terminology note:}
Throughout this paper, “augmentation’’ refers to \textit{phenomenological modeling of lidar degradations with physically interpretable parameters}, distinct from augmentation techniques used for neural network training.

\section{Methodology}
\label{sec:meth}

\setlength{\parskip}{0pt}
\setlength{\parindent}{1em}

\subsection{Framework Overview}
\label{subsec:framework_overview}

\begin{figure}[!t]
    \centering
    \includegraphics[width=0.48\textwidth]{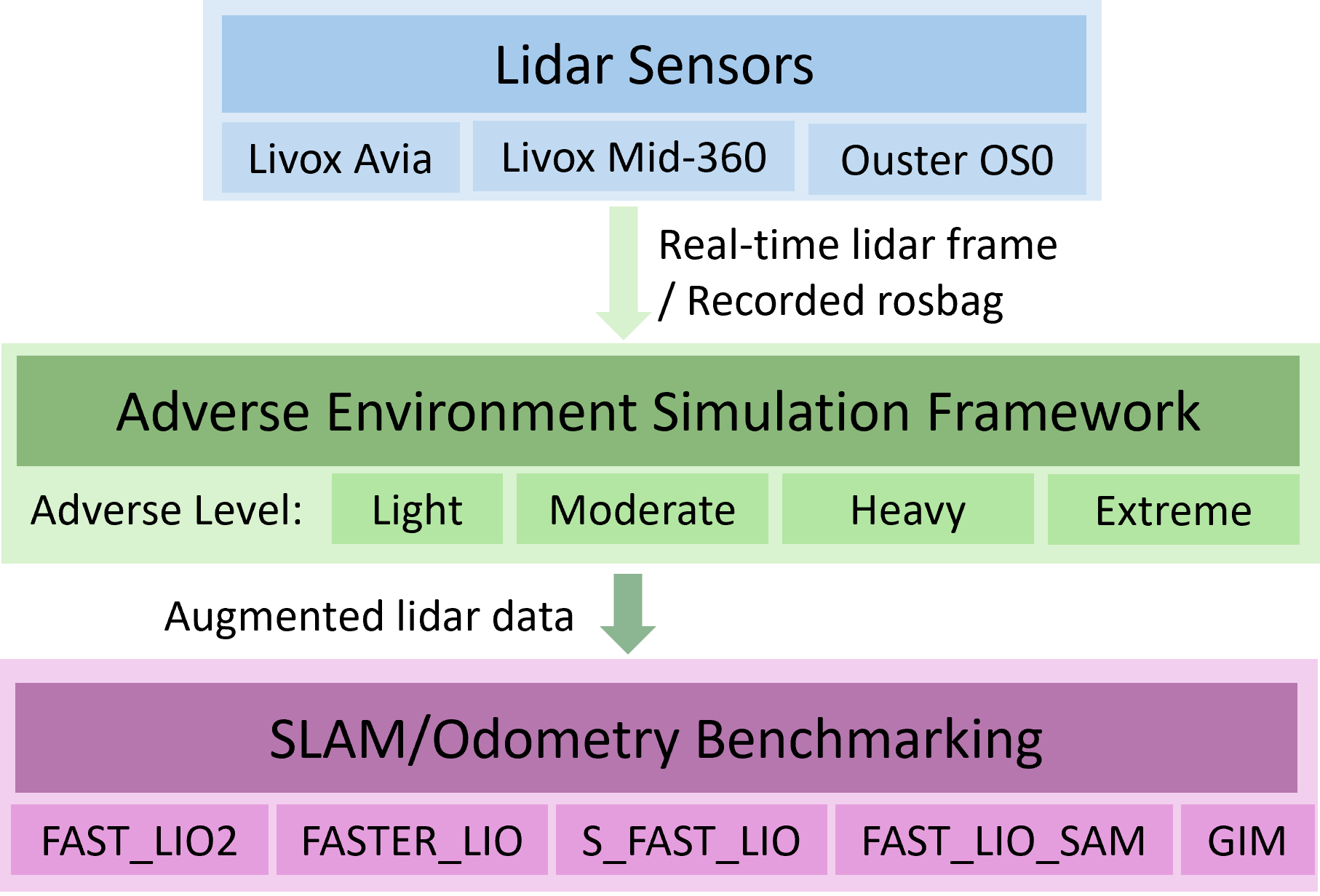}
    \caption{Architecture of the proposed adverse environment simulation framework for robustness evaluation of SLAM algorithms.}
    \label{fig:framework_diagram}
    \vspace{-1.6em}
\end{figure}

\noindent
Fig.~\ref{fig:framework_diagram} shows the architecture of the proposed sensor-aware phenomenological degradation framework. 
The system processes multi-sensor lidar streams, operating on both recorded ROS bags and real-time inputs. Incoming point clouds are parsed, timestamp-aligned, and converted into a unified representation before passing through the degradation pipeline.

Physically interpretable augmentation modules, including point dropout, occlusion, noise injection, motion distortion, and FoV reduction, are applied according to severity levels defined in configuration files. Multiple degradation types can be chained or randomized to emulate realistic adverse conditions while preserving temporal consistency using IMU timestamps when available.

The augmented point clouds are republished on dedicated ROS topics, enabling seamless integration with a wide range of LIO and SLAM systems. The framework adopts a modular architecture with real-time C++ processing and optional Python utilities for visualization and monitoring. Detailed implementation notes, including node structure and configuration examples, are provided in the open-source repository.

\subsection{Autonomous Sensor Configuration}
\label{subsec:auto_config}

\noindent
The framework includes an autonomous configuration mechanism that removes the need for manual setup across different lidar platforms. 
Sensor characteristics are inferred directly from incoming point cloud fields, allowing automatic identification of each lidar stream. 
Active topics are discovered at runtime, and corresponding augmented outputs are published using consistent naming conventions. 
Coordinate frame identifiers are preserved throughout the pipeline, ensuring compatibility with existing TF structures and downstream SLAM modules. 
This zero-configuration design simplifies deployment and prevents misconfiguration when processing heterogeneous or multi-sensor datasets.

\subsection{Scenario Design and Parameterization}
\label{subsec:scenario_design}

\noindent
Degradation scenarios are defined through configuration files that specify physically interpretable parameters for five perturbation types: dropout, field-of-view (FoV) reduction, additive noise, occlusion masking, and motion distortion. 
Parameters are grouped into four severity tiers (\textit{light}–\textit{extreme}), providing a reproducible and progressive stress-testing protocol for SLAM evaluation across diverse lidar architectures. 
Complete configuration templates and detailed parameter values are provided in the open-source repository.

\paragraph*{Design Philosophy}
The framework follows two principles: 
(i)~\textbf{physical relevance}, using parameters that reflect measurable sensor or environmental effects; and 
(ii)~\textbf{comparability}, applying identical severity tiers across sensors and experiments. 
This structured severity design aligns with recent corruption benchmarks such as SemanticKITTI-C and nuScenes-C~\cite{10286105,10205484,DBLP:journals/corr/abs-2409-10824}, which use discrete intensity levels to analyze algorithmic failure modes.

\paragraph*{Modeling Approach}
We adopt a phenomenological modeling strategy derived from observable lidar behavior rather than full radiometric or atmospheric simulation. 
The degradation modules preserve geometric consistency, temporal ordering, and each sensor's characteristic scanning pattern while enabling real-time computation suitable for both bag replay and live operation.

\paragraph*{Parameter Justification}
Severity levels span realistic ranges informed by empirical observation and sensor-physics analysis. 
Dropout levels vary from mild to severe sparsification (approximately 15–40\%), consistent with signal-loss mechanisms such as beam divergence and atmospheric attenuation~\cite{alma9914854225606531}. 
Gaussian noise levels (on the order of 1–5\,cm) reflect the typical range-measurement uncertainty observed in both solid-state and mechanical lidars under nominal and adverse conditions~\cite{9157107}. 
FoV reductions (roughly 10–40°) emulate partial angular occlusions caused by sensor housing or nearby structures, while occlusion patches remove contiguous local clusters representative of pedestrians, vegetation, or vehicle edges. 
Motion distortion is applied using realistic linear and angular velocities with per-point temporal offsets, providing physically meaningful perturbations that significantly affect geometric alignment without compromising reproducibility.

\paragraph*{Scope and Impact}
The framework focuses on dominant degradation mechanisms that influence SLAM drift and registration stability, prioritizing generality and interpretability over exhaustive environmental simulation. 
This design supports standardized robustness evaluation, cross-sensor comparison, and controlled ablation studies, providing a consistent foundation for benchmarking lidar-based SLAM systems under adverse conditions.
\subsection{Augmentation (Simulation) Modules}
\label{subsec:modules}

\noindent
The degradation modules implement the phenomenological parameters defined in the scenario configuration. 
Let a raw scan be $\mathcal{P}=\{(\mathbf{p}_i,\mathbf{a}_i,t_i)\}_{i=1}^{N}$, where $\mathbf{p}_i\in\mathbb{R}^3$ is a 3D point, $\mathbf{a}_i$ its attributes, and $t_i$ its timestamp. 
The motion model follows the continuous-time formulation widely used in lidar odometry systems such as LOAM~\cite{Zhang-RSS-14}, and all transformations preserve the spatial structure and temporal ordering of each scan.

\vspace{-0.3em}
\begin{modlist}
\item \textbf{Point Dropout.}
Sparsification is applied through a Bernoulli mask:
\[
m_i \sim \mathrm{Bernoulli}(1-r), \qquad
\mathbf{p}_i' =
\begin{cases}
\mathbf{p}_i, & m_i=1,\\[-3pt]
\varnothing, & m_i=0,
\end{cases}
\]
where $r$ is the dropout ratio.  
Structured variants remove contiguous angular or spatial regions to emulate realistic occluders.

\item \textbf{Occlusion and FoV Reduction.}
FoV cropping constrains points to angular limits in azimuth $\theta_i$ and elevation $\phi_i$, computed as
\[
\theta_i=\mathrm{atan2}(y_i,x_i), \qquad
\phi_i=\mathrm{asin}(z_i/\|\mathbf{p}_i\|),
\]
with cropping applied through bounds $|\theta_i|\le\Theta_{\max}$ and $|\phi_i|\le\Phi_{\max}$.  
Random occlusion removes $K$ spherical patches of radius $s$, retaining points satisfying
\[
\|\mathbf{p}_i - \mathbf{c}_k\| > s \quad \forall k,
\]
where occlusion centers $\mathbf{c}_k$ are uniformly sampled within a bounded range.

\item \textbf{Noise Injection.}
Gaussian and outlier noise perturb each point:
\[
\mathbf{p}_i' = \mathbf{p}_i + \boldsymbol{\epsilon}_i + o_i\boldsymbol{\eta}_i,
\]
where $\boldsymbol{\epsilon}_i \!\sim\! \mathcal{N}(0,\sigma^2\mathbf{I})$, 
$o_i\!\sim\!\mathrm{Bernoulli}(q)$, and  
$\boldsymbol{\eta}_i \!\sim\! \mathcal{N}(0,\sigma_o^2\mathbf{I})$.

\item \textbf{Motion Distortion.}
Assuming constant linear velocity $\mathbf{v}$ and angular velocity $\boldsymbol{\omega}$ over a scan:
\[
\mathbf{p}_i' = \mathbf{R}(\Delta t_i)\mathbf{p}_i + \mathbf{v}\Delta t_i,
\qquad
\Delta t_i = t_i - t_0.
\]
The rotation is computed using the SE(3) exponential map, with a first-order approximation
\[
\mathbf{R}(\Delta t_i) \approx \mathbf{I} + [\boldsymbol{\omega}\Delta t_i]_\times
\]
applied for small angles $\|\boldsymbol{\omega}\Delta t_i\| < 0.1$~\cite{lynch2017modernrobotics}.

\item \textbf{Scan Sparsification.}
Systematic subsampling retains every $s$-th point:
\[
\mathcal{P}'=\{(\mathbf{p}_i,\mathbf{a}_i,t_i)\mid i\equiv 0\pmod{s}\}.
\]
\end{modlist}

\vspace{-0.5em}

\subsection{Integration and Reproducibility}
\label{subsec:integration}

\noindent
All modules are parameterized through configuration files with severity tiers (\textit{light}–\textit{extreme}), with optional runtime overrides via ROS parameters. 
Sensor-specific message conversions ensure compatibility with downstream SLAM systems, and IMU measurements are interpolated to estimate velocities used in motion distortion. 
Deterministic module chaining, fixed random seeds, and diagnostic logging provide full reproducibility across experiments.


\subsection{Implementation and Experimental Setup}
\label{subsec:implementation_setup}

\noindent
The framework is implemented as a hybrid C++/Python ROS package, with time-critical augmentation and I/O handled in C++ for real-time performance, and Python utilities supporting visualization and interactive parameter tuning. Although validated primarily on recorded rosbag data, the ROS-native design supports direct integration with live lidar streams.

\noindent
Experiments were conducted using the multi-sensor sequences from our publicly available dataset \href{https://github.com/TIERS/multi_modal_lidar_dataset.git}{GitHub repository}~\cite{11248862}, which provides synchronized Livox Avia, Livox Mid-360, and Ouster OS0-128 point clouds across indoor and outdoor environments. The system was tested on a workstation running Ubuntu~20.04 with an AMD Ryzen~9~7945HX CPU and 32\,GB RAM. Degradation settings were specified in YAML configuration files that define four severity tiers (\textit{light}–\textit{extreme}), which are automatically applied at runtime by the autonomous sensor configuration module (Section~\ref{subsec:auto_config}).To assess robustness under controlled degradations, five representative lidar odometry and SLAM systems, FAST-LIO2~\cite{9697912}, FASTER-LIO~\cite{9718203}, S-FAST-LIO~\cite{zlwang7_sfastlio_2023}, FAST-LIO-SAM~\cite{engcang_fastliosam_2023}, and GLIM~\cite{KOIDE2024104750}, were evaluated. These algorithms cover direct, iterative, and mapping-based approaches, enabling a broad assessment of robustness across SLAM paradigms.
\section{Experimental Results}\label{sec:exper_results}
\begin{table*}[t]
\centering
\caption{APE (mean $\pm$ std, meters) across severity levels (L=Light, M=Moderate, H=Heavy, X=Extreme). Columns are grouped by sensor (Avia, Mid360, Ouster).}
\label{tab:slam_results_stacked}
\resizebox{\textwidth}{!}{
\begin{tabular}{l|
cccc!{\color{gray!60}\vrule width 0.4pt}
cccc!{\color{gray!60}\vrule width 0.4pt}
cccc}
\toprule
\multicolumn{13}{c}{\textbf{\textit{IndoorOffice1}}} \\
\midrule
\multirow{2}{*}{\textbf{SLAM}} &
\multicolumn{4}{c!{\color{gray!60}\vrule width 0.4pt}}{\textbf{Avia\_10Hz}} &
\multicolumn{4}{c!{\color{gray!60}\vrule width 0.4pt}}{\textbf{Mid360\_10Hz}} &
\multicolumn{4}{c}{\textbf{Ouster\_10Hz}} \\
& L & M & H & X & L & M & H & X & L & M & H & X \\
\midrule
FAST-LIO2     & \textbf{0.161$\pm$0.075} & 0.263$\pm$0.152 & 0.445$\pm$0.310 & \textbf{0.407$\pm$0.285} & 0.0468$\pm$0.0180 & 0.0453$\pm$0.0214 & 0.0499$\pm$\textbf{0.0234} & 0.0534$\pm$\textbf{0.0259} & \textbf{0.0643}$\pm$0.0442 & 0.0739$\pm$0.0495 & 0.0816$\pm$0.0558 & \textbf{0.152}$\pm$0.072 \\
FASTER-LIO    & 0.191$\pm$0.076 & 0.227$\pm$0.106 & \textbf{0.300$\pm$0.210} & 4.059$\pm$1.070 & 0.0918$\pm$0.0435 & 0.0988$\pm$0.0476 & 0.105$\pm$0.0480 & 0.0897$\pm$0.0345 & 0.0809$\pm$\textbf{0.0410} & 0.0953$\pm$\textbf{0.0456} & 0.126$\pm$0.0536 & 0.180$\pm$\textbf{0.069} \\
S-FAST-LIO    & 0.441$\pm$0.081 & 0.679$\pm$0.313 & 0.585$\pm$0.419 & 6.249$\pm$1.666 & \textbf{0.0412}$\pm$0.0198 & \textbf{0.0415$\pm$0.0205} & 0.0465$\pm$0.0248 & \textbf{0.0448}$\pm$0.0265 & 0.0760$\pm$0.0532 & 0.0759$\pm$0.0521 & \textbf{0.0810}$\pm$0.0547 & 0.160$\pm$0.085 \\
GLIM          & 0.301$\pm$0.148 & 0.258$\pm$0.186 & 1.016$\pm$0.276 & 3.432$\pm$3.158 & 0.0427$\pm$0.0280 & 0.0428$\pm$0.0282 & \textbf{0.0454}$\pm$0.0321 & 0.0522$\pm$0.0405 & 0.0947$\pm$0.0534 & 0.0953$\pm$0.0551 & 0.0992$\pm$0.0595 & 0.170$\pm$0.092 \\
FAST-LIO-SAM  & 0.176$\pm$0.0917 & \textbf{0.167$\pm$0.0836} & 0.386$\pm$0.389 & 1.479$\pm$0.524 & 0.0460$\pm$\textbf{0.0172} & 0.0476$\pm$0.0209 & 0.0503$\pm$0.0248 & 0.0518$\pm$0.0261 & 0.0650$\pm$0.0436 & \textbf{0.0732}$\pm$0.0485 & 0.0821$\pm$\textbf{0.0524} & 0.163$\pm$0.0757 \\
\bottomrule
\toprule
\multicolumn{13}{c}{\textbf{\textit{IndoorOffice2}}} \\
\midrule
FAST-LIO2     & 0.1047$\pm$0.0842 & 0.648$\pm$0.428 & 1.600$\pm$0.457 & 4.245$\pm$1.340 & 0.0422$\pm$0.0302 & 0.0458$\pm$0.0280 & 0.0493$\pm$0.0267 & 0.0596$\pm$0.0350 & \textbf{0.0602}$\pm$0.0592 & 0.0692$\pm$0.0638 & \textbf{0.0724}$\pm$0.0589 & 0.171$\pm$0.080 \\
FASTER-LIO    & \textbf{0.0885}$\pm$0.0645 & \textbf{0.128$\pm$0.0701} & 1.144$\pm$\textbf{0.312} & 3.631$\pm$1.262 & \textbf{0.0358}$\pm$0.0275 & \textbf{0.0376}$\pm$0.0301 & \textbf{0.0388$\pm$0.0264} & \textbf{0.0423$\pm$0.0233} & 0.0724$\pm$\textbf{0.0570} & 0.0829$\pm$0.0613 & 0.0915$\pm$\textbf{0.0563} & 0.177$\pm$\textbf{0.076} \\
S-FAST-LIO    & 0.110$\pm$\textbf{0.0543} & 0.905$\pm$0.419 & 1.515$\pm$0.791 & 7.011$\pm$1.921 & 0.0414$\pm$\textbf{0.0273} & 0.0394$\pm$\textbf{0.0272} & 0.0432$\pm$0.0289 & 0.0563$\pm$0.0330 & 0.0697$\pm$0.0671 & \textbf{0.0678$\pm$0.0609} & 0.0765$\pm$0.0638 & \textbf{0.143}$\pm$0.078 \\
GLIM          & 0.218$\pm$0.149 & 0.653$\pm$0.498 & \textbf{1.123}$\pm$0.499 & 4.169$\pm$2.393 & 0.0786$\pm$0.0908 & 0.0862$\pm$0.1224 & 0.0764$\pm$0.0464 & 0.0790$\pm$0.0891 & 0.0879$\pm$0.0722 & 0.0897$\pm$0.0772 & 0.0874$\pm$0.0749 & 0.163$\pm$0.086 \\
FAST-LIO-SAM  & 0.0950$\pm$0.0667 & 0.363$\pm$0.237 & 1.659$\pm$0.395 & \textbf{1.144$\pm$0.828} & 0.0405$\pm$0.0295 & 0.0397$\pm$0.0298 & 0.0490$\pm$0.0314 & 0.0570$\pm$0.0314 & 0.0638$\pm$0.0630 & 0.0730$\pm$0.0624 & 0.0745$\pm$0.0657 & 0.171$\pm$0.076 \\
\bottomrule
\toprule
\multicolumn{13}{c}{\textbf{\textit{OutdoorRoad}}} \\
\midrule
FAST-LIO2     & \textbf{0.302}$\pm$0.147 & \textbf{0.420$\pm$0.189} & 0.623$\pm$0.280 & 0.544$\pm$0.276 & 0.385$\pm$0.181 & 0.445$\pm$0.197 & 0.516$\pm$0.223 & 0.454$\pm$0.248 & 0.603$\pm$0.310 & 0.622$\pm$0.320 & 0.615$\pm$0.291 & \textbf{0.423$\pm$0.190} \\
FASTER-LIO    & 0.338$\pm$\textbf{0.122} & 0.513$\pm$0.292 & 0.469$\pm$\textbf{0.212} & \textbf{0.481$\pm$0.233} & \textbf{0.312$\pm$0.152} & \textbf{0.400$\pm$0.173} & \textbf{0.511$\pm$0.195} & 0.496$\pm$\textbf{0.227} & \textbf{0.449$\pm$0.228} & \textbf{0.434$\pm$0.224} & \textbf{0.367}$\pm$0.207 & 0.690$\pm$0.335 \\
S-FAST-LIO    & 0.346$\pm$0.137 & 0.492$\pm$0.236 & \textbf{0.468}$\pm$0.237 & 0.690$\pm$0.521 & 0.320$\pm$0.156 & 0.457$\pm$0.192 & 0.728$\pm$0.325 & \textbf{0.425}$\pm$0.260 & 0.644$\pm$0.345 & 0.537$\pm$0.277 & 0.374$\pm$\textbf{0.168} & 0.980$\pm$0.350 \\
GLIM          & 0.802$\pm$0.372 & 1.701$\pm$0.519 & 5.171$\pm$1.096 & 4.178$\pm$1.200 & 0.591$\pm$0.451 & 0.905$\pm$0.842 & 0.936$\pm$0.824 & 1.840$\pm$1.454 & 0.836$\pm$0.435 & 0.860$\pm$0.529 & 1.305$\pm$0.821 & 1.381$\pm$0.663 \\
FAST-LIO-SAM  & 0.338$\pm$0.149 & 0.427$\pm$0.240 & 0.474$\pm$0.231 & 0.822$\pm$0.469 & 0.369$\pm$0.170 & 0.499$\pm$0.205 & 0.594$\pm$0.237 & 0.546$\pm$0.273 & 0.611$\pm$0.327 & 0.632$\pm$0.328 & 0.405$\pm$0.190 & 0.624$\pm$0.281 \\
\bottomrule
\end{tabular}}
\vspace{-2.0em}
\end{table*}

The implementation of the augmentation framework delivers not only the augmentation methodologies themselves but also a real-time visualization dashboard, as described in detail in Section~\ref{subsec:technical-impl}. The average processing time for augmentation operations is typically well below 20\,ms with the computer device mentioned in the above sections.
\subsection{Quantitative SLAM Evaluation}
\label{subsec:quantitative}

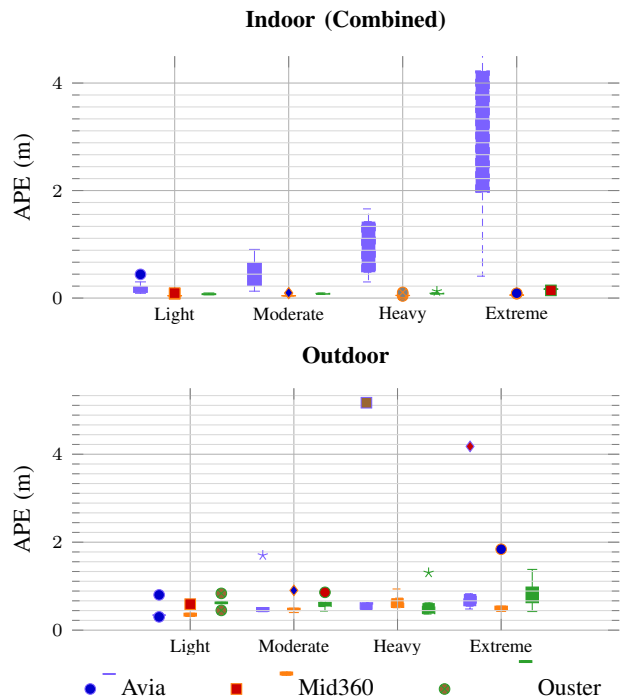
\begin{figure}[!t]
    \centering
    \def\plotwidth{0.45\textwidth}
    \def\plotheight{4.8cm}
\definecolor{AviaC}{HTML}{7B61FF}
\definecolor{Mid360C}{HTML}{FF7F0E}
\definecolor{OusterC}{HTML}{2CA02C}

\begin{tikzpicture}
\begin{groupplot}[
  group style={
    group size=1 by 2, 
    vertical sep=1.2cm,
    xlabels at=edge bottom,
  },
  width=\columnwidth,              
  height=4.8cm,                    
  ylabel={APE (m)},
  yminorgrids=true,
  major grid style={gray!60},
  minor grid style={gray!30},
  axis on top,
  tick label style={font=\scriptsize},
  label style={font=\small},
  boxplot/draw direction=y,
  boxplot/box extend=0.12,
  enlarge x limits=0.10,
]

\nextgroupplot[
  title={Indoor (Combined)},
  title style={font=\small\bfseries},
  xtick={1.5,2.5,3.5,4.5},
  xticklabels={Light, Moderate, Heavy, Extreme},
  axis line style={draw=none},
  xmin=1.0, xmax=5.0,
  ymin=0, ymax=4.5,
  yminorgrids=true,        
  minor y tick num=8,      
  grid=both,               
  xlabel={},
]

\addplot+[boxplot, fill=AviaC, draw=AviaC, boxplot/draw position=1.2]
table[row sep=\\, y index=0] {
data\\
0.161\\0.191\\0.441\\0.301\\0.176\\
0.1047\\0.0885\\0.110\\0.218\\0.0950\\
};
\addplot+[boxplot, fill=Mid360C, draw=Mid360C, boxplot/draw position=1.5]
table[row sep=\\, y index=0] {
data\\
0.0468\\0.0918\\0.0412\\0.0427\\0.0460\\
0.0422\\0.0358\\0.0414\\0.0786\\0.0405\\
};
\addplot+[boxplot, fill=OusterC, draw=OusterC, boxplot/draw position=1.8]
table[row sep=\\, y index=0] {
data\\
0.0643\\0.0809\\0.0760\\0.0947\\0.0650\\
0.0602\\0.0724\\0.0697\\0.0879\\0.0638\\
};

\addplot+[boxplot, fill=AviaC, draw=AviaC, boxplot/draw position=2.2]
table[row sep=\\, y index=0] {
data\\
0.263\\0.227\\0.679\\0.258\\0.167\\
0.648\\0.128\\0.905\\0.653\\0.363\\
};
\addplot+[boxplot, fill=Mid360C, draw=Mid360C, boxplot/draw position=2.5]
table[row sep=\\, y index=0] {
data\\
0.0453\\0.0988\\0.0415\\0.0428\\0.0476\\
0.0458\\0.0376\\0.0394\\0.0862\\0.0397\\
};
\addplot+[boxplot, fill=OusterC, draw=OusterC, boxplot/draw position=2.8]
table[row sep=\\, y index=0] {
data\\
0.0739\\0.0953\\0.0759\\0.0953\\0.0732\\
0.0692\\0.0829\\0.0678\\0.0897\\0.0730\\
};

\addplot+[boxplot, fill=AviaC, draw=AviaC, boxplot/draw position=3.2]
table[row sep=\\, y index=0] {
data\\
0.445\\0.300\\0.585\\1.016\\0.386\\
1.600\\1.144\\1.515\\1.123\\1.659\\
};
\addplot+[boxplot, fill=Mid360C, draw=Mid360C, boxplot/draw position=3.5]
table[row sep=\\, y index=0] {
data\\
0.0499\\0.105\\0.0465\\0.0454\\0.0503\\
0.0493\\0.0388\\0.0432\\0.0764\\0.0490\\
};
\addplot+[boxplot, fill=OusterC, draw=OusterC, boxplot/draw position=3.8]
table[row sep=\\, y index=0] {
data\\
0.0816\\0.126\\0.0810\\0.0992\\0.0821\\
0.0724\\0.0915\\0.0765\\0.0874\\0.0745\\
};

\addplot+[boxplot, fill=AviaC, draw=AviaC, boxplot/draw position=4.2]
table[row sep=\\, y index=0] {
data\\
0.407\\4.059\\6.249\\3.432\\1.479\\
4.245\\3.631\\7.011\\4.169\\1.144\\
};
\addplot+[boxplot, fill=Mid360C, draw=Mid360C, boxplot/draw position=4.5]
table[row sep=\\, y index=0] {
data\\
0.0534\\0.0897\\0.0448\\0.0522\\0.0518\\
0.0596\\0.0423\\0.0563\\0.0790\\0.0570\\
};
\addplot+[boxplot, fill=OusterC, draw=OusterC, boxplot/draw position=4.8]
table[row sep=\\, y index=0] {
data\\
0.152\\0.180\\0.160\\0.170\\0.163\\
0.171\\0.177\\0.143\\0.163\\0.171\\
};

\nextgroupplot[
  title={Outdoor},
  title style={font=\small\bfseries},
  xtick={1.5,2.5,3.5,4.5},
  xticklabels={Light, Moderate, Heavy, Extreme},
  axis line style={draw=none},
  xmin=0.8, xmax=5.2,
  ymin=0, ymax=5.5,
  yminorgrids=true,        
  minor y tick num=8,      
  grid=both,               
  xlabel={},
  legend style={
    at={(0.5,-0.10)},           
    anchor=north,
    legend columns=3, 
    draw=none, 
    font=\small,
    /tikz/every even column/.append style={column sep=0.8cm}
  },
]

\addplot+[boxplot, fill=AviaC, draw=AviaC, boxplot/draw position=1.2]
table[row sep=\\, y index=0] {
data\\
0.302\\0.338\\0.346\\0.802\\0.338\\
};
\addplot+[boxplot, fill=Mid360C, draw=Mid360C, boxplot/draw position=1.5]
table[row sep=\\, y index=0] {
data\\
0.385\\0.312\\0.320\\0.591\\0.369\\
};
\addplot+[boxplot, fill=OusterC, draw=OusterC, boxplot/draw position=1.8]
table[row sep=\\, y index=0] {
data\\
0.603\\0.449\\0.644\\0.836\\0.611\\
};

\addplot+[boxplot, fill=AviaC, draw=AviaC, boxplot/draw position=2.2]
table[row sep=\\, y index=0] {
data\\
0.420\\0.513\\0.492\\1.701\\0.427\\
};
\addplot+[boxplot, fill=Mid360C, draw=Mid360C, boxplot/draw position=2.5]
table[row sep=\\, y index=0] {
data\\
0.445\\0.400\\0.457\\0.905\\0.499\\
};
\addplot+[boxplot, fill=OusterC, draw=OusterC, boxplot/draw position=2.8]
table[row sep=\\, y index=0] {
data\\
0.622\\0.434\\0.537\\0.860\\0.632\\
};

\addplot+[boxplot, fill=AviaC, draw=AviaC, boxplot/draw position=3.2]
table[row sep=\\, y index=0] {
data\\
0.623\\0.469\\0.468\\5.171\\0.474\\
};
\addplot+[boxplot, fill=Mid360C, draw=Mid360C, boxplot/draw position=3.5]
table[row sep=\\, y index=0] {
data\\
0.516\\0.511\\0.728\\0.936\\0.594\\
};
\addplot+[boxplot, fill=OusterC, draw=OusterC, boxplot/draw position=3.8]
table[row sep=\\, y index=0] {
data\\
0.615\\0.367\\0.374\\1.305\\0.405\\
};

\addplot+[boxplot, fill=AviaC, draw=AviaC, boxplot/draw position=4.2]
table[row sep=\\, y index=0] {
data\\
0.544\\0.481\\0.690\\4.178\\0.822\\
};
\addplot+[boxplot, fill=Mid360C, draw=Mid360C, boxplot/draw position=4.5]
table[row sep=\\, y index=0] {
data\\
0.454\\0.496\\0.425\\1.840\\0.546\\
};
\addplot+[boxplot, fill=OusterC, draw=OusterC, boxplot/draw position=4.8]
table[row sep=\\, y index=0] {
data\\
0.423\\0.690\\0.980\\1.381\\0.624\\
};

\legend{Avia, Mid360, Ouster}

\end{groupplot}
\end{tikzpicture}
    \caption{
   Distribution of Absolute Pose Error (APE) across degradation severities (\textit{light--extreme}) for three lidar sensors (Avia, Mid360, Ouster) in both indoor and outdoor settings. Each box aggregates APE values across multiple SLAM back-ends, revealing cross-sensor robustness trends and sensitivity to degradation strength.
   }
    \label{fig:boxplot}
    \vspace{-1.0em}
\end{figure}

\noindent
Quantitative robustness analysis was performed using the \texttt{evo\_ape} tool\footnote{https://github.com/MichaelGrupp/evo} to compute Absolute Pose Error (APE) across all severity tiers. Fig.~\ref{fig:boxplot} summarizes APE distributions for three lidar types in both indoor and outdoor environments. Performance remains stable under \textit{light} and \textit{moderate} degradations, while errors and variance increase progressively toward the \textit{extreme} setting. 
Livox Mid-360 exhibits the lowest sensitivity to degradation, consistent across all SLAM back-ends. Overall, the results demonstrate that the proposed phenomenological degradations induce controlled, interpretable stress conditions that allow systematic benchmarking of lidar-based SLAM algorithms.

\subsection{Discussion}
\label{subsec:discussion}

\noindent
\textbf{Framework Positioning and Design Trade-offs:}
Compared with existing corruption benchmarks (SemanticKITTI-C, nuScenes-C~\cite{10205484,10286105}) that project 2D perturbations into point clouds, or simulation engines such as CARLA~\cite{Dosovitskiy17} and lidarsim~\cite{9157601} requiring full scene geometry, our phenomenological approach operates directly on real lidar scans. This preserves authentic noise characteristics, temporal patterns, and sensor-specific artifacts while enabling real-time processing (less than 20\,ms per frame). Although atmospheric phenomena (fog density, rain attenuation) are not explicitly modeled, this design choice ensures cross-sensor generality and practical in-field robustness evaluation.

\noindent
\textbf{Sensor-Specific Robustness Patterns:}
Table~\ref{tab:slam_results_stacked} and Fig.~\ref{fig:boxplot} reveal distinct resilience profiles across lidar architectures. Livox Mid-360 exhibits the strongest robustness, maintaining sub-decimeter APE across all severities in indoor conditions; its non-repetitive scanning pattern and uniform spatial coverage preserve feature distributions under dropout and occlusion. Livox Avia shows a \emph{threshold-driven} failure mode: performance remains stable under light/moderate degradation but deteriorates sharply at extreme levels (e.g., 0.16\,m $\rightarrow$ 4--7\,m), indicating sensitivity to density collapse. Ouster OS0-128 presents intermediate behavior with higher variance in extreme indoor settings, suggesting that structured scanning patterns are more vulnerable to systematic occlusion.

\noindent
\textbf{Algorithm–Sensor Interaction Effects:}
FAST-LIO2 and S-FAST-LIO maintain strong median accuracy but occasionally produce large-error outliers (4--7\,m) when dropout exceeds $\sim$35\%, indicating re-initialization failures rather than gradual drift. GLIM yields higher baseline error but lower variance, implying that its global scan-to-map consistency is more tolerant to sparsity. FASTER-LIO demonstrates the most uniform behavior across sensors. 

\noindent
\textbf{Environmental Context Dependency:}
Indoor and outdoor conditions exhibit fundamentally different degradation responses. Indoor environments produce higher failure rates at extreme severity (Avia: $8\times$; Mid-360: $1.2\times$), reflecting the reduced feature redundancy in confined spaces. Outdoor sequences degrade more gradually (Mid-360: $\sim1.5\times$ increase) due to richer geometric structure but exhibit higher absolute APE because of reduced density at long range. These results emphasize that robustness is not scale-invariant; algorithms tuned for outdoor navigation may fail abruptly in degraded indoor environments.

\noindent
\textbf{Practical Implications:}
The findings provide actionable guidance for system design. 
(i)~Mid-360 is well-suited for safety-critical indoor autonomy requiring high degradation resilience, whereas Avia offers cost efficiency in controlled outdoor domains. 
(ii)~The presence of catastrophic outliers motivates degradation-aware monitoring (point count, FoV coverage, feature density) to trigger fallback modes before divergence. 
(iii)~Complementary failure modes suggest heterogeneous sensor fusion strategies, where Mid-360 handles degraded conditions and Avia provides efficient nominal coverage. 
(iv)~For worst-case robustness, GLIM's predictable error profile may be advantageous, while FASTER-LIO's consistency suits real-time embedded platforms.
\begin{figure*}[t]
    \centering
    \begin{subfigure}{0.24\textwidth}
        \centering
       \includegraphics[width=0.99\textwidth]{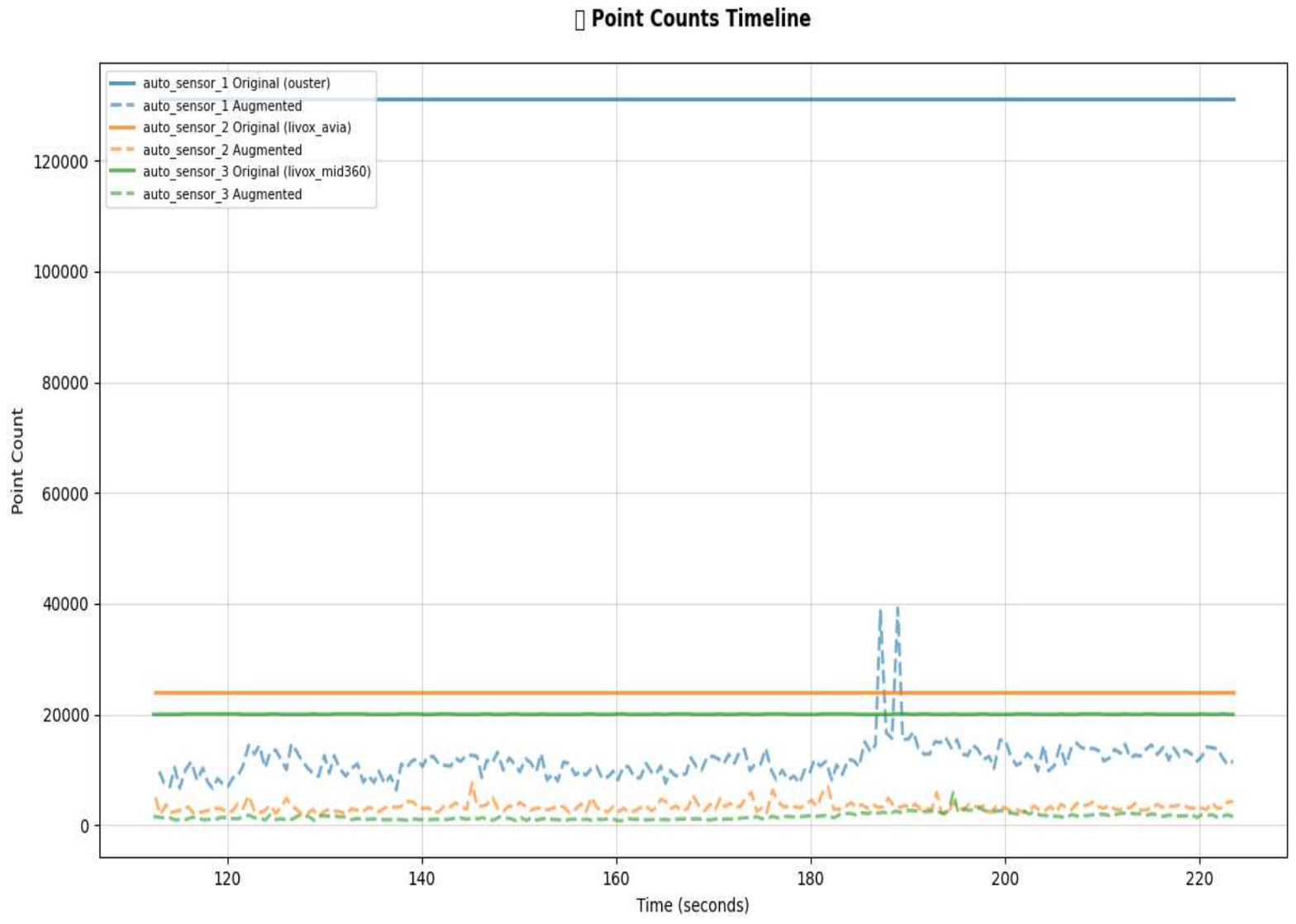}
    \end{subfigure}
        \begin{subfigure}{0.24\textwidth}
        \centering
       \includegraphics[width=0.99\textwidth]{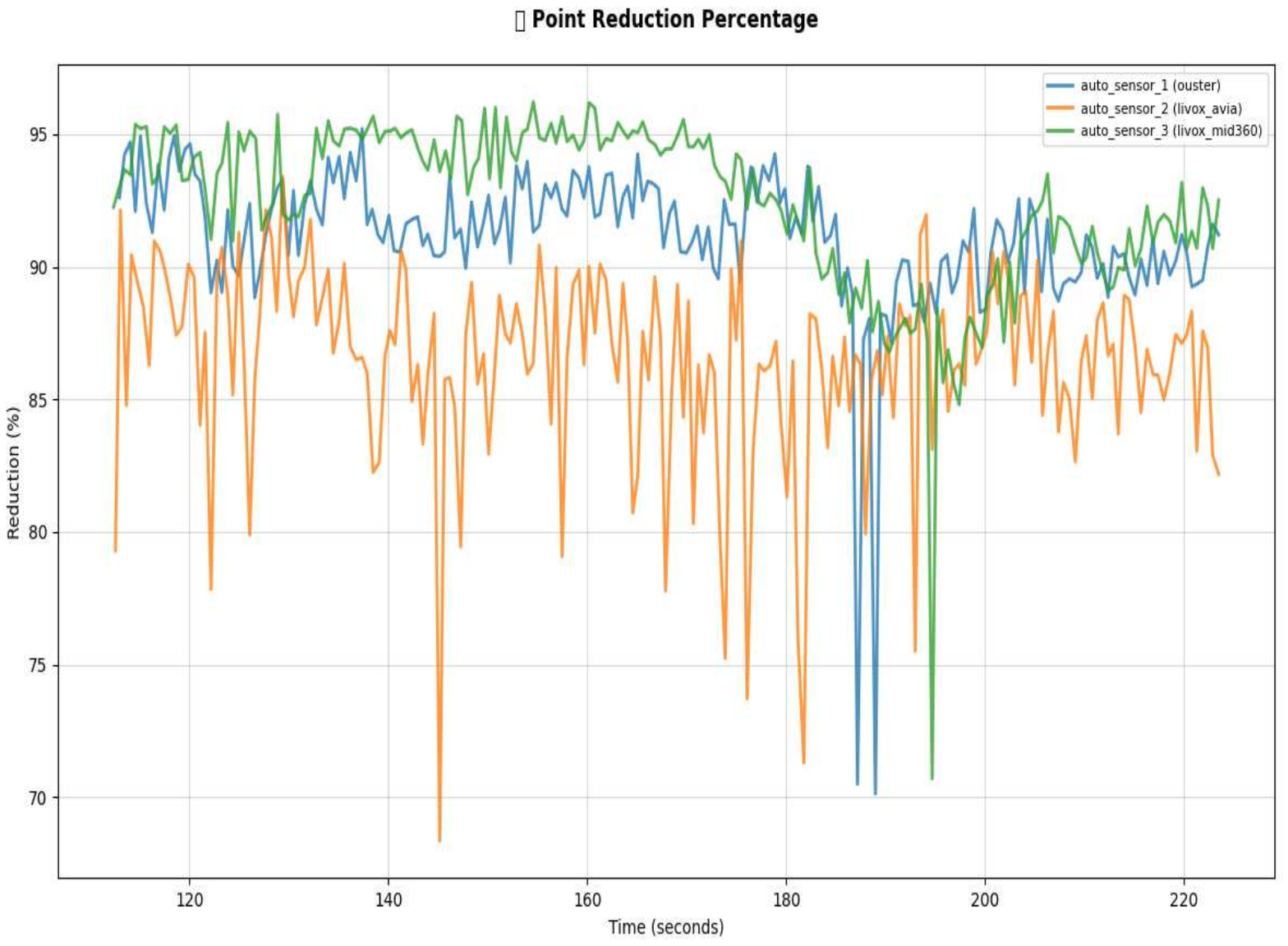}
    \end{subfigure}
    \hfill
        \begin{subfigure}{0.24\textwidth}
        \centering
       \includegraphics[width=0.99\textwidth]{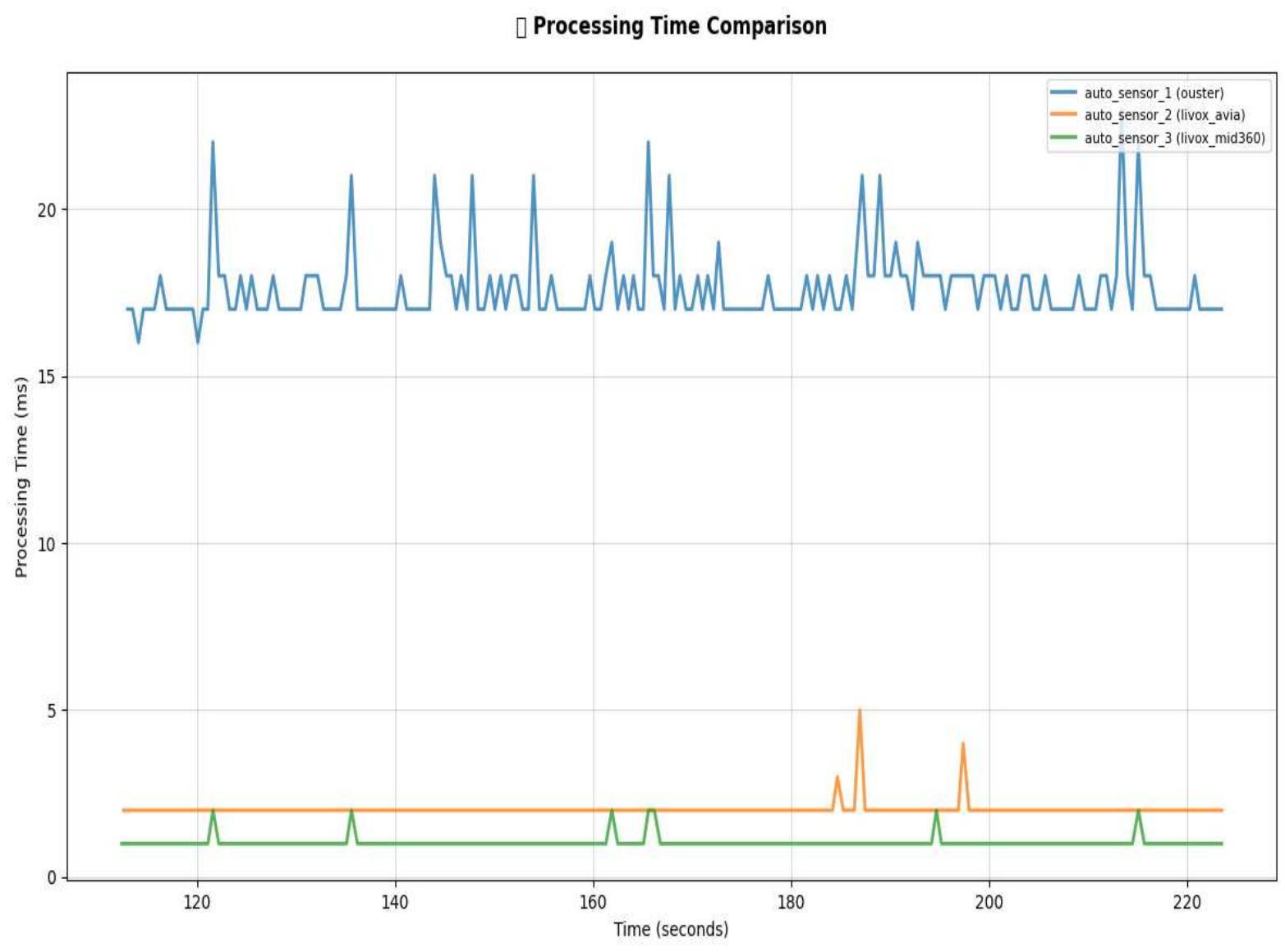}
    \end{subfigure}
        \begin{subfigure}{0.24\textwidth}
        \centering
       \includegraphics[width=0.99\textwidth]{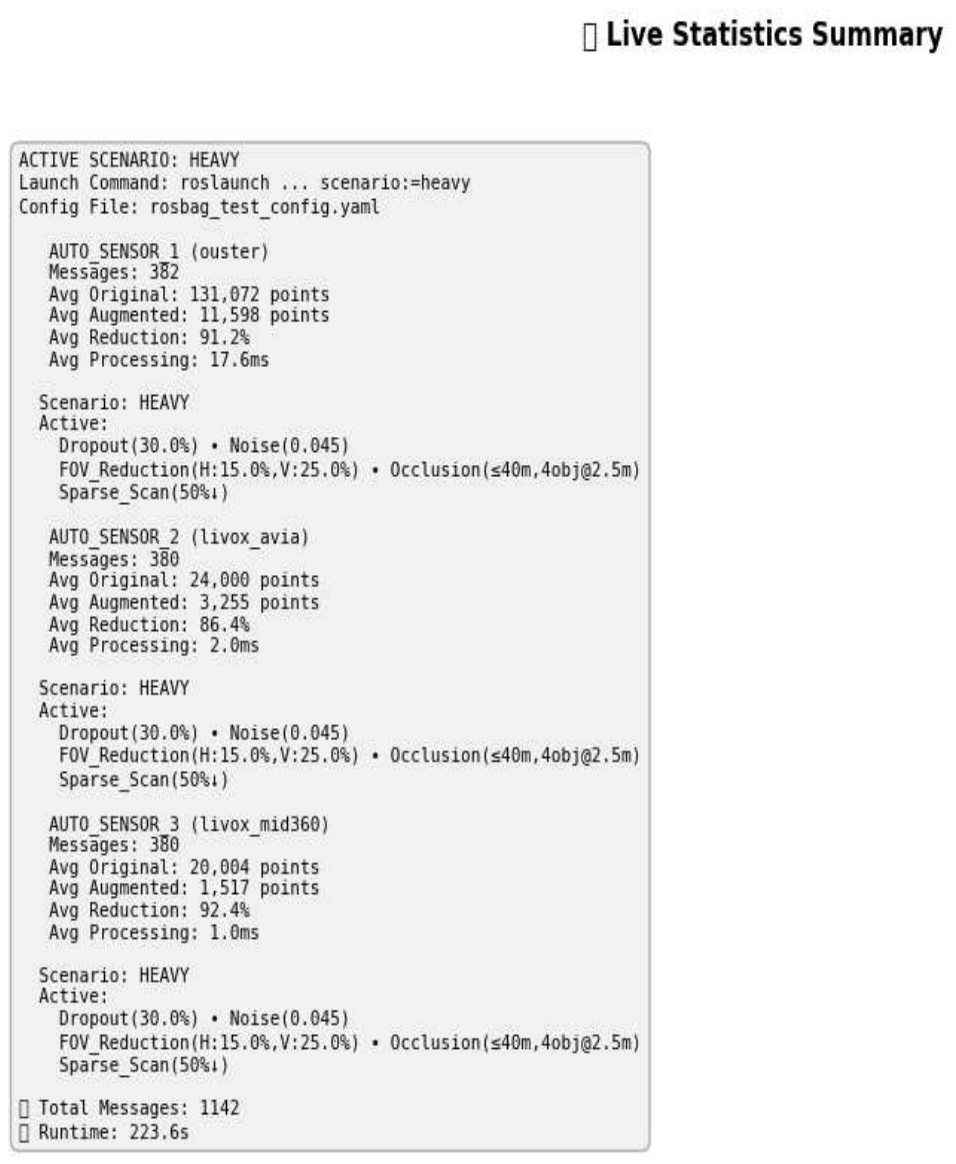}
    \end{subfigure}
    \caption{Real-time augmentation dashboard for the \textit{OutdoorRoad} dataset under the \textit{heavy} scenario, illustrating per-sensor point counts, reduction ratios, and processing-time comparisons. Each lidar (Ouster, Livox~Avia, Livox~Mid-360) is automatically detected and processed concurrently through the autonomous configuration module.}
    \label{fig:dashboard}
    \vspace{-10pt}
\end{figure*}

\section{Conclusion}\label{sec:conclusion}
\noindent
We presented a sensor-aware phenomenological framework for lidar degradation, enabling systematic and reproducible robustness evaluation of SLAM algorithms under adverse conditions. 
Unlike generic corruption benchmarks or simulation engines requiring detailed scene geometry, our approach operates directly on raw lidar measurements with real-time performance ($<20$\,ms per frame), while preserving hardware-specific scan patterns and temporal consistency. 
The framework integrates autonomous topic and sensor detection with modular degradation models, dropout, FoV reduction, noise, occlusion, and motion distortion, organized into four severity tiers (\textit{light}–\textit{extreme}) to provide physically interpretable, cross-sensor stress testing.

Experimental validation across three lidar architectures and five SLAM systems revealed distinct robustness behaviors shaped by both sensor design and environmental context. Mid-360 consistently exhibited strong resilience, Avia showed threshold-driven failures under extreme degradation, and Ouster displayed intermediate sensitivity shaped by its structured scanning pattern. 
These results demonstrate that the proposed framework provides a practical and generalizable tool for evaluating lidar-based SLAM performance, enabling the community to analyze robustness trends, identify failure modes, and design degradation-aware perception systems for real-world deployment.

\newpage
\section{Technical Implementation Details}\label{subsec:technical-impl}


The implementation of the proposed framework is publicly available\footnote{\url{https://github.com/mawuto/lidar_augmentation_cpp_ws}} and follows a hybrid C++/Python design. Time-critical augmentation, sensor parsing, and ROS I/O are implemented in C++ for real-time performance, while Python utilities support BEV visualization, online statistics, and diagnostic logging. The system operates on both rosbag files and live lidar streams without modification.

\subsection{Configuration and Severity Tiers}
All degradation parameters are defined in the repository under 
\texttt{config/rosbag\_test\_config.yaml}.  
Four severity tiers, \textit{light}, \textit{moderate}, \textit{heavy}, and \textit{extreme}, map to the phenomenological models in Section~\ref{subsec:modules} via structured namespaces 
(e.g., \texttt{augmentations.dropout.*},  
\texttt{augmentations.noise.*},  
\texttt{augmentations.fov.*}).  
These settings control dropout ratios, FoV limits, noise levels, occlusion masks, motion distortion velocities, and sparsification, and can be overridden at runtime using ROS parameters.

\subsection{Integration and Reproducibility}
\label{subsec:integration}
The augmentation node automatically discovers available lidar and IMU topics and applies sensor-specific parsing rules: Ouster fields (\texttt{x,y,z, intensity, reflectivity, line}) are processed natively, while Livox data are converted between \texttt{PointCloud2} and \texttt{livox\_ros\_driver/CustomMsg} when required by downstream SLAM systems. IMU buffers are synchronized using trapezoidal integration to estimate per-point velocities for motion distortion.

Deterministic module ordering, fixed random seeds, and logged configuration states ensure complete reproducibility. A minimal reproduction pipeline, which launches the augmentation node with a selected scenario and replays rosbag files, is documented in the repository, along with utilities for visualization and APE evaluation.

\subsection{Real-Time Visualization and Performance}\label{subsec:dashboard}
Fig.~\ref{fig:dashboard} shows the real-time monitoring dashboard under the \textit{heavy} scenario. The framework processes multiple sensors concurrently with augmentation times below 20\,ms per frame (Ouster: 19.8\,ms; Avia: 5.4\,ms; Mid-360: 4.1\,ms), ensuring compatibility with 10\,Hz operation. The autonomous configuration module eliminates manual setup, enabling scalable multi-sensor deployments in both offline and real-time settings.

\subsection{Example Original and  Augmented Point Cloud }
\label{subsec:qualitative}


\begin{figure}[h]
    \centering
    \begin{subfigure}{0.42\textwidth}
        \includegraphics[width=0.99\textwidth,trim=10 20 10 20, clip]{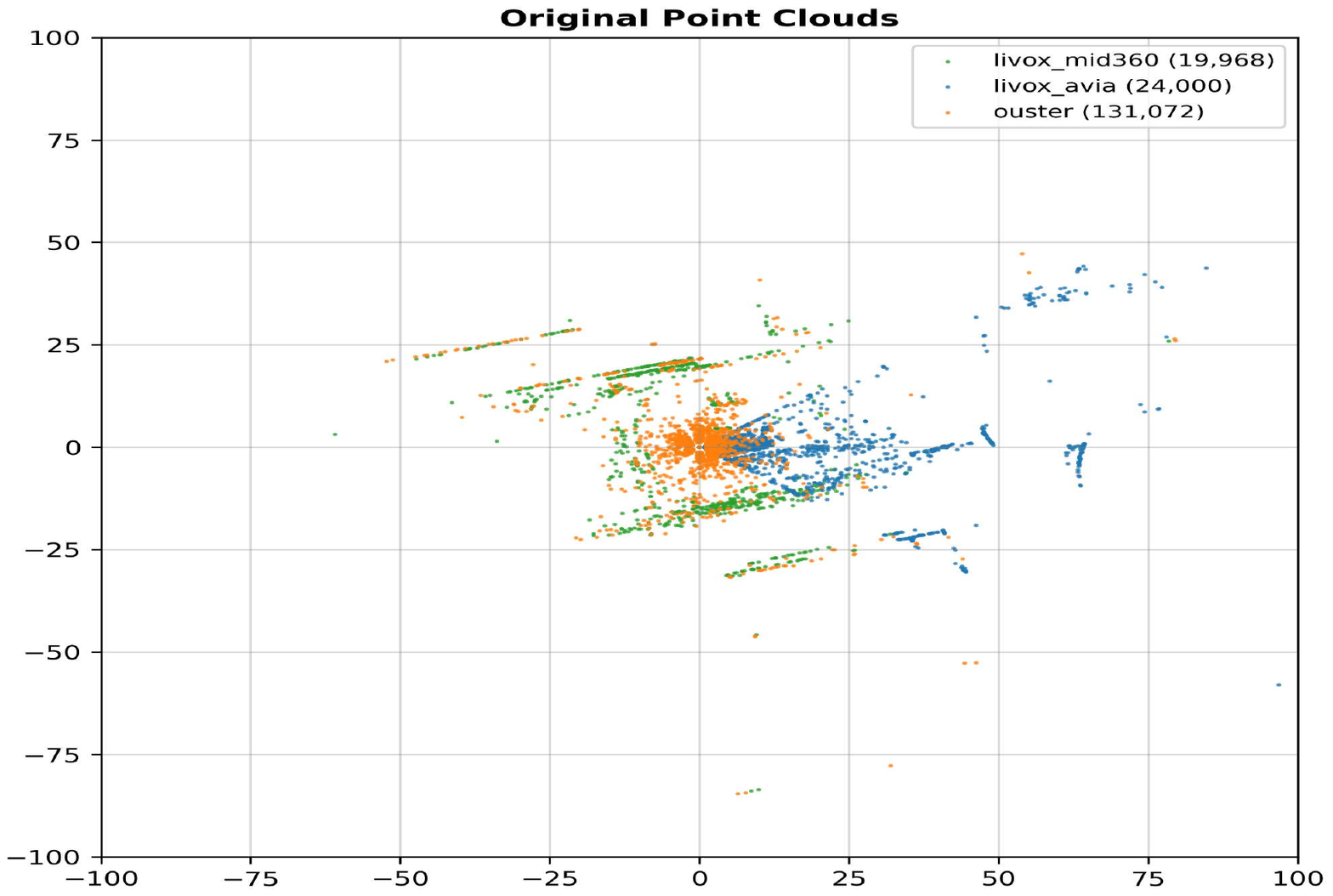}   
    \end{subfigure}
    \hfill
    \begin{subfigure}{0.42\textwidth}
        \includegraphics[width=0.99\textwidth,trim=10 20 10 20, clip]{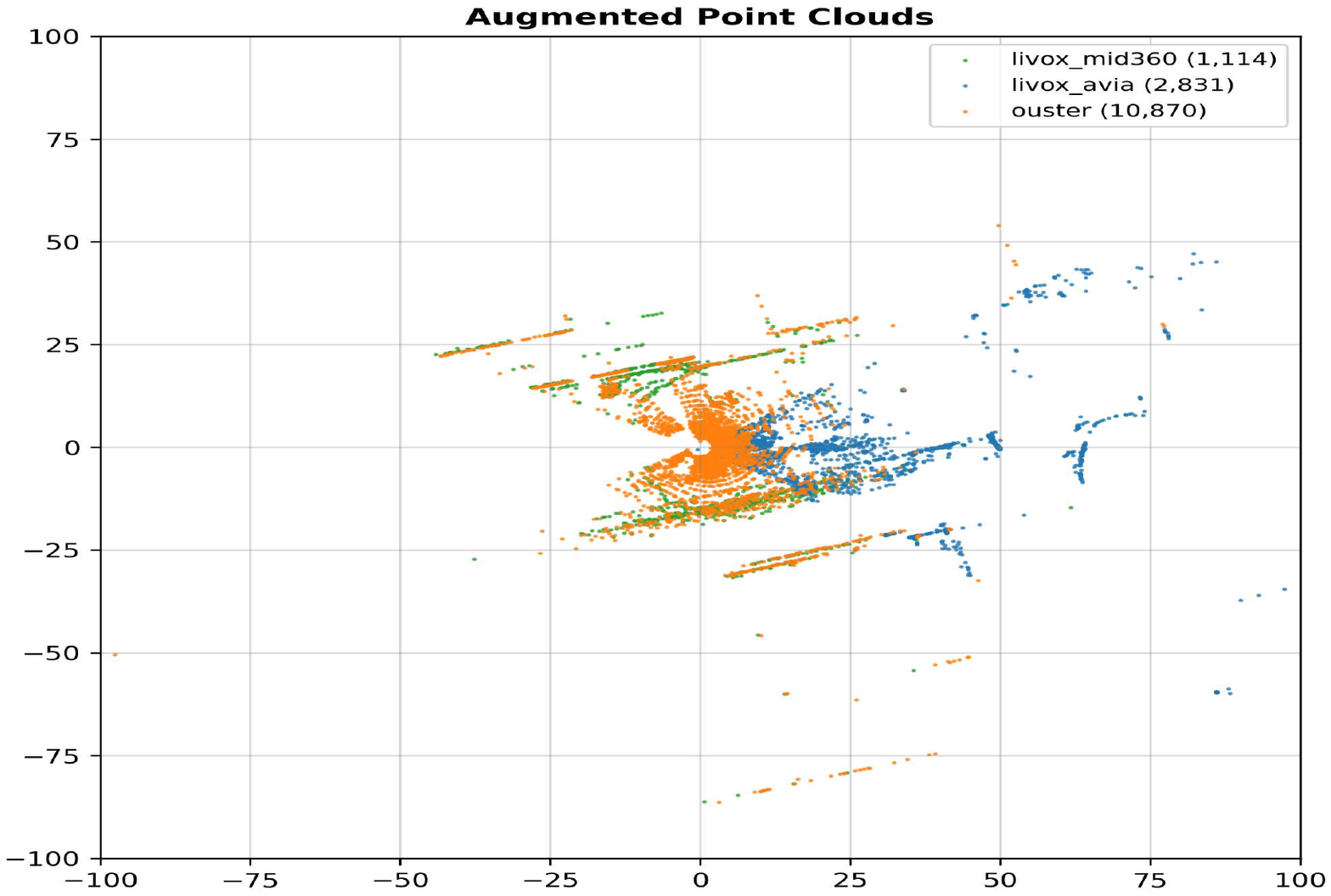}   
    \end{subfigure}
    \caption{
    Bird’s-eye-view (BEV) visualization of multi-lidar frames from the \textit{OutdoorRoad} dataset under the \textit{heavy} degradation scenario. The framework preserves cross-sensor alignment and structural geometry while introducing realistic dropout and occlusion consistent with outdoor environments.}
    \label{fig:bev}
\end{figure}


Fig.~\ref{fig:bev} compares original and augmented BEV representations under the \textit{heavy} scenario. 
The augmented frames exhibit realistic sparsification, partial occlusion, and sectoral loss while retaining global geometry and cross-sensor alignment. These effects emulate real outdoor conditions, including vegetation, shadowed regions, and reflective surfaces. The results confirm that the proposed framework produces physically grounded degradations without disrupting spatial calibration or temporal consistency required by SLAM pipelines.



\bibliographystyle{unsrt}
\bibliography{bibliography}

@article{pritzl2025degradation,
  title={Degradation-Aware Cooperative Multi-Modal GNSS-Denied Localization Leveraging LiDAR-Based Robot Detections},
  author={Pritzl, V{\'a}clav and Yu, Xianjia and Westerlund, Tomi and {\v{S}}t{\v{e}}p{\'a}n, Petr and Saska, Martin},
  journal={arXiv preprint arXiv:2510.20480},
  year={2025}
}

@article{sier2023benchmark,
  title={A benchmark for multi-modal lidar slam with ground truth in gnss-denied environments},
  author={Sier, Ha and Li, Qingqing and Yu, Xianjia and Pe{\~n}a Queralta, Jorge and Zou, Zhuo and Westerlund, Tomi},
  journal={Remote Sensing},
  volume={15},
  number={13},
  pages={3314},
  year={2023},
  publisher={MDPI}
}

@inproceedings{xiao2022polarmix,
  title     = {PolarMix: A General Data Augmentation Technique for LiDAR Point Clouds},
  author    = {Xiao, Aoran and Huang, Jiaxing and Guan, Dayan and Cui, Kaiwen and Lu, Shijian and Shao, Ling},
  booktitle = {Advances in Neural Information Processing Systems},
  volume    = {35},
  pages     = {21605--21618},
  year      = {2022}
}

@inproceedings{sebek2023real3daug,
  title     = {Real3D-Aug: Point Cloud Augmentation by Placing Real Objects with Occlusion Handling for 3D Detection and Segmentation},
  author    = {Šebek, Petr and Pokorný, Šimon and Vacek, Patrik and Svoboda, Tomáš},
  booktitle = {Computer Vision Winter Workshop (CVWW) 2023},
  year      = {2023}
}

@article{ZHANG2023146,
  title = {Perception and sensing for autonomous vehicles under adverse weather conditions: A survey},
  journal = {ISPRS Journal of Photogrammetry and Remote Sensing},
  volume = {196},
  pages = {146-177},
  year = {2023},
  issn = {0924-2716},
  doi = {10.1016/j.isprsjprs.2022.12.021},
  url = {https://www.sciencedirect.com/science/article/pii/S0924271622003367},
  author = {Yuxiao Zhang and Alexander Carballo and Hanting Yang and Kazuya Takeda}
}

@ARTICLE{10898027,
  author={Wu, Qi and Chen, Xieyuanli and Xu, Xiangyu and Zhong, Xinliang and Qu, Xingwei and Xia, Songpengcheng and Liu, Guoqing and Liu, Liu and Yu, Wenxian and Pei, Ling},
  journal={IEEE Transactions on Instrumentation and Measurement}, 
  title={UA-LIO: An Uncertainty-Aware LiDAR-Inertial Odometry for Autonomous Driving in Urban Environments}, 
  year={2025},
  volume={74},
  number={},
  pages={1-12},
  doi={10.1109/TIM.2025.3544286}}

@INPROCEEDINGS{9157601,
  author={Manivasagam, Sivabalan and Wang, Shenlong and Wong, Kelvin and Zeng, Wenyuan and Sazanovich, Mikita and Tan, Shuhan and Yang, Bin and Ma, Wei-Chiu and Urtasun, Raquel},
  booktitle={2020 IEEE/CVF Conference on Computer Vision and Pattern Recognition (CVPR)}, 
  title={LiDARsim: Realistic LiDAR Simulation by Leveraging the Real World}, 
  year={2020},
  volume={},
  number={},
  pages={11164-11173},
  doi={10.1109/CVPR42600.2020.01118}}

@article{DBLP:journals/corr/abs-2409-10824,
  author       = {Bo Yang and
                  Tri Minh Triet Pham and
                  Jinqiu Yang},
  title        = {Evaluating and Improving the Robustness of LiDAR-based Localization
                  and Mapping},
  journal      = {CoRR},
  volume       = {abs/2409.10824},
  year         = {2024},
  url          = {https://doi.org/10.48550/arXiv.2409.10824},
  doi          = {10.48550/ARXIV.2409.10824},
  eprinttype    = {arXiv},
  eprint       = {2409.10824},
  bibsource    = {dblp computer science bibliography, https://dblp.org}
}

@INPROCEEDINGS{10186539,
  author={Dreissig, Mariella and Scheuble, Dominik and Piewak, Florian and Boedecker, Joschka},
  booktitle={2023 IEEE Intelligent Vehicles Symposium (IV)}, 
  title={Survey on LiDAR Perception in Adverse Weather Conditions}, 
  year={2023},
  volume={},
  number={},
  pages={1-8},
  doi={10.1109/IV55152.2023.10186539}}

@INPROCEEDINGS{10205484,
  author={Dong, Yinpeng and Kang, Caixin and Zhang, Jinlai and Zhu, Zijian and Wang, Yikai and Yang, Xiao and Su, Hang and Wei, Xingxing and Zhu, Jun},
  booktitle={2023 IEEE/CVF Conference on Computer Vision and Pattern Recognition (CVPR)}, 
  title={Benchmarking Robustness of 3D Object Detection to Common Corruptions in Autonomous Driving}, 
  year={2023},
  volume={},
  number={},
  pages={1022-1032},
  doi={10.1109/CVPR52729.2023.00105}}

@ARTICLE{10286105,
  author={Li, Shuangzhi and Wang, Zhijie and Juefei-Xu, Felix and Guo, Qing and Li, Xingyu and Ma, Lei},
  journal={IEEE Transactions on Multimedia}, 
  title={Common Corruption Robustness of Point Cloud Detectors: Benchmark and Enhancement}, 
  year={2025},
  volume={27},
  number={},
  pages={848-859},
  doi={10.1109/TMM.2023.3318317}}

@misc{zhan2023realaugrealisticscenesynthesis,
      title={Real-Aug: Realistic Scene Synthesis for LiDAR Augmentation in 3D Object Detection}, 
      author={Jinglin Zhan and Tiejun Liu and Rengang Li and Jingwei Zhang and Zhaoxiang Zhang and Yuntao Chen},
      year={2023},
      eprint={2305.12853},
      archivePrefix={arXiv},
      primaryClass={cs.CV},
      url={https://arxiv.org/abs/2305.12853}, 
}

@INPROCEEDINGS{Zhang-RSS-14, 
    AUTHOR    = {Ji Zhang AND Sanjiv Singh}, 
    TITLE     = {LOAM: Lidar Odometry and Mapping in Real-time}, 
    BOOKTITLE = {Proceedings of Robotics: Science and Systems}, 
    YEAR      = {2014}, 
    ADDRESS   = {Berkeley, USA}, 
    MONTH     = {July},
    DOI       = {10.15607/RSS.2014.X.007} 
}

@book{alma9914854225606531,
author = {Weitkamp, Claus},
address = {New York, NY},
booktitle = {Lidar : Range-Resolved Optical Remote Sensing of the Atmosphere},
edition = {1st ed. 2005.},
isbn = {1-280-60948-6},
language = {eng},
publisher = {Springer New York},
series = {Springer Series in Optical Sciences, 102},
title = {Lidar : Range-Resolved Optical Remote Sensing of the Atmosphere },
year = {2005},
}

@INPROCEEDINGS{9157107,
  author={Bijelic, Mario and Gruber, Tobias and Mannan, Fahim and Kraus, Florian and Ritter, Werner and Dietmayer, Klaus and Heide, Felix},
  booktitle={2020 IEEE/CVF Conference on Computer Vision and Pattern Recognition (CVPR)}, 
  title={Seeing Through Fog Without Seeing Fog: Deep Multimodal Sensor Fusion in Unseen Adverse Weather}, 
  year={2020},
  volume={},
  number={},
  pages={11679-11689},
  doi={10.1109/CVPR42600.2020.01170}}

@book{lynch2017modernrobotics,
  title     = {Modern Robotics: Mechanics, Planning, and Control},
  author    = {Lynch, Kevin M. and Park, Frank C.},
  publisher = {Cambridge University Press},
  year      = {2017},
  isbn      = {9781107156302},
  note      = {1st Edition}
}

@ARTICLE{9718203,
  author={Bai, Chunge and Xiao, Tao and Chen, Yajie and Wang, Haoqian and Zhang, Fang and Gao, Xiang},
  journal={IEEE Robotics and Automation Letters}, 
  title={Faster-LIO: Lightweight Tightly Coupled Lidar-Inertial Odometry Using Parallel Sparse Incremental Voxels}, 
  year={2022},
  volume={7},
  number={2},
  pages={4861-4868},
  doi={10.1109/LRA.2022.3152830}}

@ARTICLE{9697912,
  author={Xu, Wei and Cai, Yixi and He, Dongjiao and Lin, Jiarong and Zhang, Fu},
  journal={IEEE Transactions on Robotics}, 
  title={FAST-LIO2: Fast Direct LiDAR-Inertial Odometry}, 
  year={2022},
  volume={38},
  number={4},
  pages={2053-2073},
  doi={10.1109/TRO.2022.3141876}}

@article{KOIDE2024104750,
title = {GLIM: 3D range-inertial localization and mapping with GPU-accelerated scan matching factors},
journal = {Robotics and Autonomous Systems},
volume = {179},
pages = {104750},
year = {2024},
issn = {0921-8890},
doi = {https://doi.org/10.1016/j.robot.2024.104750},
url = {https://www.sciencedirect.com/science/article/pii/S0921889024001349},
author = {Kenji Koide and Masashi Yokozuka and Shuji Oishi and Atsuhiko Banno}
}

@misc{zlwang7_sfastlio_2023,
  author       = {Wang, Z.},
  title        = {S-FAST-LIO: A Simplified Implementation of FAST-LIO},
  year         = {2023},
  howpublished = {\url{https://github.com/zlwang7/S-FAST_LIO}},
  note         = {Accessed: 2025-11-27}
}

@misc{engcang_fastliosam_2023,
  author       = {Engcang},
  title        = {FAST-LIO-SAM: A SLAM Implementation Combining FAST-LIO2 with Pose Graph Optimization and Loop Closing Based on LIO-SAM},
  year         = {2023},
  howpublished = {\url{https://github.com/engcang/FAST-LIO-SAM}},
  note         = {Accessed: 2025-11-27}
}

@ARTICLE{11248862,
  author={Felix, Doumegna Mawuto Koudjo and Yu, Xianjia and Zhang, Jiaqiang and Ha, Sier and Zou, Zhuo and Westerlund, Tomi},
  journal={IEEE Robotics and Automation Letters}, 
  title={Understanding Lidar Variability: A Dataset and Comparative Study Featuring Dome-Shaped, Solid-State, and Spinning Lidars}, 
  year={2026},
  volume={11},
  number={1},
  pages={570-577},
  doi={10.1109/LRA.2025.3632749}}

@inproceedings{Dosovitskiy17,
  title = {{CARLA}: {An} Open Urban Driving Simulator},
  author = {Alexey Dosovitskiy and German Ros and Felipe Codevilla and Antonio Lopez and Vladlen Koltun},
  booktitle = {Proceedings of the 1st Annual Conference on Robot Learning},
  pages = {1--16},
  year = {2017}
}

\end{document}